\documentclass[letterpaper, 10 pt, conference]{IEEEtran}
\usepackage[letterpaper, left=0.75in, right=0.75in, top=0.75in, bottom=0.76in]{geometry}
\IEEEoverridecommandlockouts
\usepackage{amsmath,amssymb,amsfonts}
\usepackage{algorithmic}
\usepackage{booktabs}
\usepackage{graphicx}
\usepackage[caption=false,font=footnotesize]{subfig}
\usepackage[numbers,sort&compress]{natbib}
\usepackage{tabularx}
\usepackage{textcomp}
\usepackage{xcolor}
\usepackage{hyperref}

\begin{document}
\bstctlcite{BSTcontrol}

\newcommand{\SAPS}{\textcolor[RGB]{255,180,100}{\textbf{SAPS}}}

\title{\vspace{0.24in}\SAPS{}{\textcolor[RGB]{255,180,100}:} Shared Autonomy for Policy Steering by Blending Teleoperation with a Pretrained VLA\\}

\newcommand{\pizerofive}{\(\pi_{0.5}\)}

\author{
\IEEEauthorblockN{
Crystal Zhou\textsuperscript{*},
Jehan Yang\textsuperscript{*},
Douglas J. Weber\textsuperscript{$\dagger$},
and Zackory Erickson\textsuperscript{$\dagger$}
}
\IEEEauthorblockA{
Carnegie Mellon University\\
Pittsburgh, PA, USA\\
crystalz@andrew.cmu.edu,
jehan@cmu.edu
}
\thanks{\(^*\)These authors contributed equally to this work; authorship order was determined by a coin flip. Both are the corresponding authors. \(^\dagger\)These authors advised equally for this work. This material is based upon work supported by the National Science Foundation under Grant No. 2341352.}

}

\maketitle

\begin{abstract}

Vision-Language-Action (VLA) models perform well in generalist robot manipulation but remain brittle to out-of-distribution perturbations, while full human teleoperation is cognitively demanding. To bridge this gap, we introduce Shared Autonomy for Policy Steering (SAPS), a lightweight framework that blends real-time human teleoperation commands with pretrained VLA actions. SAPS operates purely at the action level, requiring no policy retraining, auxiliary dynamics models, or architectural modifications. We evaluate three arbitration strategies across simulated (LIBERO, CALVIN) and real-world evaluations. In a 14-person user study, SAPS achieved up to 77.4\% hardware success, compared with 8.3\% for full autonomy and 34.5\% for teleoperation. Furthermore, active human intervention was required only 45\% of the time, reducing NASA-TLX workload by 20.8\% in simulation and 28.2\% on hardware. A case study with an expert user with quadriplegia similarly demonstrated matched or improved success while reducing completion time. These results highlight SAPS as a highly effective, model-agnostic approach for reliable real-world deployment and assistive teleoperation. A website with additional details is located at shared-autonomy-policy.github.io.

\end{abstract}


\section{Introduction}
Recent Vision-Language-Action (VLA) models have accelerated robotic manipulation by enabling generalist policies trained on large datasets to perform diverse tasks~\citep{pi05,gr00tn1,openvla,black2024pi0,zitkovich2023rt2}. Despite strong zero-shot capabilities and benefits from multimodal language labels~\citep{barreiros2025lbm,lin2026cotrainlbm}, these models remain brittle under out-of-distribution (OOD) perturbations in object placement, appearance, and task specification~\citep{zhou2025liberopro}. In these settings, autonomous policies can fail by executing incorrect tasks or stalling in inefficient recovery behaviors. 

Recent work addresses these failures through inference-time policy steering for pretrained models~\citep{wang2025itps,liu2026vls,wu2025forewarn,wu2025saysteering}. However, these methods require auxiliary dynamics models, modifications to diffusion samplers, or explicit reasoning modules, and can still fall short of human teleoperators who can recover from task failures~\citep{padmanabha2024independence,yang2026bimanualemg,hu2025racrobotlearninglonghorizon}. Pure teleoperation improves robustness, but requires continuous precise control and high user workload. This motivates shared autonomy~\citep{tao2025lams,tao2025ilsa,song2026assistron}, which combines the manipulation proficiency of foundation models with intermittent corrective input from a user. 

\begin{figure}[t]
    \centering
    \includegraphics[width=1.0\linewidth]{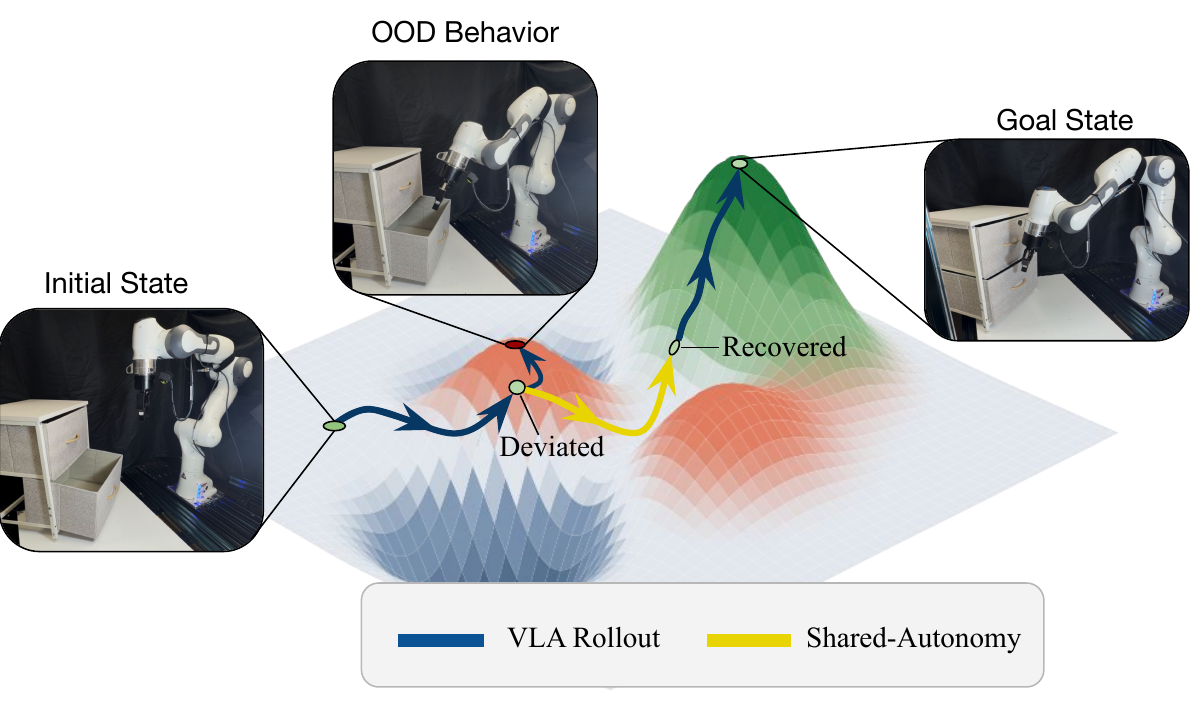}
    \vspace{-2em}
    \caption{During VLA rollout, the policy frequently executes actions that lead the policy into out-of-distribution states, or cause the policy to perform the wrong task. Through SAPS, the user guides execution toward states from which the VLA can successfully resume.}
    \label{fig:conceptual-with-screenshots}
    \vspace{-2em}
\end{figure}

We introduce SAPS (Shared Autonomy for Policy Steering), a model-agnostic framework that blends real-time human teleoperation commands with pretrained VLA actions. Unlike prior steering methods, SAPS operates post-inference at the action level, requiring no architectural modifications, auxiliary models, or policy retraining. With simple arbitration strategies, intermittent human corrections guide the policy away from or through failure states from which the pretrained policy can resume autonomous task completion, as shown in Fig.~\ref{fig:conceptual-with-screenshots}.

Our contributions are as follows:
\begin{itemize}
\item We introduce SAPS, a lightweight shared-autonomy framework that steers pretrained robot policies and VLAs by blending policy actions with human teleoperation commands, without requiring retraining, auxiliary models, or sampler modification.
\item We propose and evaluate three arbitration strategies for balancing human and VLA control: fixed blending, full takeover, and a dynamic cosine-similarity strategy that scales policy autonomy based on the geometric agreement between human and policy actions.
\item To evaluate SAPS across a variety of users, we run an able-bodied user study (\(N=14\)), an evaluation with expert users, and a case study with a user with quadriplegia. We show that SAPS significantly improves performance across LIBERO, LIBERO-PRO, CALVIN, and on a real Franka arm. 
\end{itemize}

\section{Related Works}

\subsection{Robot Foundation Models}

Recent generalist robot policies scale toward vision--language--action (VLA) models conditioned on language, vision, and proprioception. Models such as RT-2~\citep{zitkovich2023rt2}, OpenVLA~\citep{openvla}, Octo~\citep{octo_2023}, and the \(\pi_0\) family~\citep{black2024pi0,pi05} demonstrate rapid progress toward increasingly capable and generalizable robot policies. However, these models can remain brittle under visual, spatial, and language perturbations, as demonstrated by LIBERO-PRO~\citep{zhou2025liberopro}, highlighting a gap between broad autonomous capability and the flexibility required for deployment in out-of-distribution states and environments. This gap motivates our use of shared autonomy.

\subsection{Policy Steering}

Policy steering modifies a base policy's behavior at inference time to satisfy goals or constraints not fully captured during training. ITPS~\cite{wang2025itps} steers diffusion-policy sampling toward a human-specified spatial target, DynaGuide~\cite{du2025dynaguidesteeringdiffusionpolices} uses an auxiliary dynamics model to guide denoising toward likely future states, VLS~\cite{liu2026vls} uses vision-language models to synthesize trajectory-differentiable rewards, and FOREWARN~\cite{wu2025forewarn} uses predicted latent action outcomes with VLM reasoning for runtime steering. More recently, Flow Reversal Steering (FRS)~\cite{tang2026frs} transforms coarse human or VLM reference actions into actions from nearby modes of a flow-matching generalist policy by reversing the reference action through the policy's flow field. $\pi_{0.7}$~\cite{intelligence2026pi} instead incorporates inference-time verbal coaching. These approaches steer behavior within the policy's sampling, flow, or decoding process. In contrast, SAPS leaves policy inference unchanged and performs shared-autonomy arbitration only on the policy's final predicted actions.

\subsection{Shared Autonomy}

Shared autonomy considers how control and effort are distributed between a human operator and an autonomous system. Classical approaches often infer user intent and arbitrate between human and autonomous policies. Dragan and Srinivasa~\cite{dragan2013policy} formalized this interaction as policy blending, in which an arbitration function combines human and autonomous control. Javdani et al.~\cite{javdani2015shared} instead formulate assistance through goal-belief inference and expected cost-to-go minimization. Jain and Argall~\cite{jain2018recursive} use directional agreement between human and autonomous commands, measured using cosine similarity, as a signal for shared-control intent inference. Gopinath et al.~\cite{gopinath2017human} further demonstrate that preferred assistance can be user dependent: in a study with 17 participants, including four with spinal cord injury, some users favored retaining control rather than maximizing objective task performance.

Recent methods integrate learned policies and generative models into shared autonomy. LAMS~\cite{tao2025lams} uses LLM-driven control-mode switching, while ILSA~\cite{tao2025ilsa} incrementally improves assistance from user interaction and Real-to-Sim-to-Real Shared Autonomy~\cite{sha2026efficient} trains a residual copilot using a digital twin and human surrogate. To the Noise and Back~\cite{yoneda2023diffusha} transforms user actions through partial diffusion, while FlashBack~\cite{sun2025flashback} uses consistency models to make generative shared autonomy practical at real-time inference rates. Assistron~\cite{song2026assistron} applies shared autonomy to an off-the-shelf VLA, automatically identifying interaction phases where human intervention is needed and incorporating user commands through flow-matching guidance. In contrast, SAPS requires neither intent inference nor access to the VLA's generative process: it arbitrates directly between human commands and the final actions produced by the pretrained policy.

Shared autonomy is particularly important in assistive teleoperation, where the human remains part of the control loop during deployment. Prior systems have shown that shared-control assistance can enable high-DoF robotic manipulation through limited interfaces, including intracortical BCI control~\cite{muelling2017autonomy}. More recent assistive systems have incorporated foundation models into real-home robot autonomy stacks while preserving direct user control through teleoperation interfaces~\cite{padmanabha2024independence,yang2026bimanualemg}. These settings motivate evaluating not only whether assistance improves task success, but also how much continuous control and workload it places on the operator.

\section{Methodology}

\begin{figure}[t]
    \centering
    \includegraphics[width=0.8\linewidth]{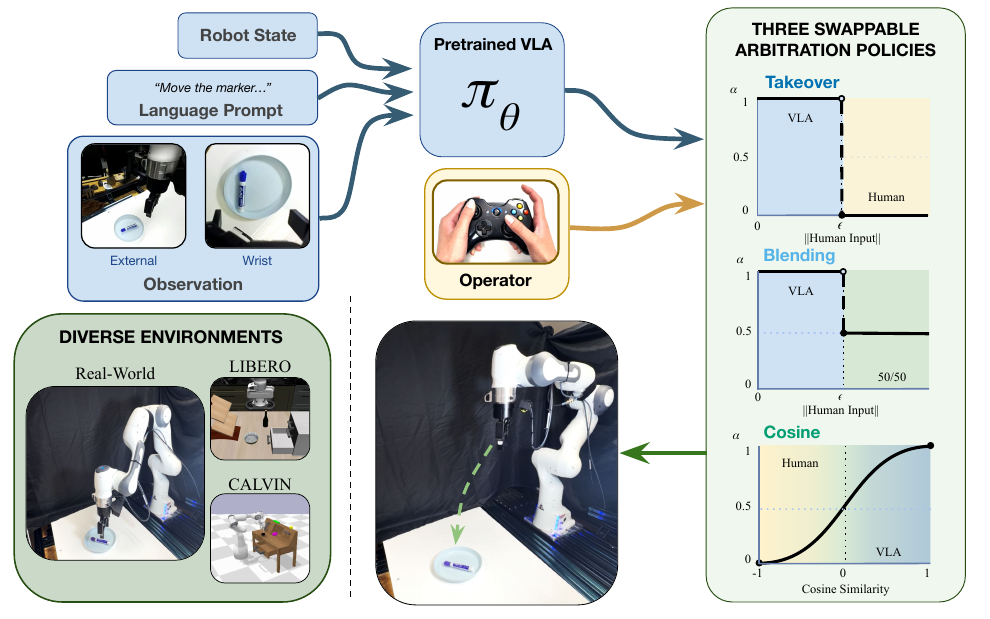}
    \vspace{-1em}
    \caption{At every control step, the VLA backbone $\pi_{\theta}$
    takes in the camera frames, robot state, and language prompt to
    propose an autonomous action.
    In parallel, the operator may supply
    a teleoperation action. We compare three VLA-operator arbitration methods that produce $\alpha$:
    a Takeover arbitrator, a Blending arbitrator outputting a 50/50 blend, and a Cosine-similarity-weighted sigmoid driven by the two actions.}
    \label{fig:fig_1}
    \vspace{-2em}
\end{figure}

\subsection{Policies}
Experiments use pretrained \pizerofive{} Vision-Language-Action (VLA) policies finetuned on the LIBERO, CALVIN, and DROID datasets. For CALVIN, to compare with previous baselines in \citet{du2025dynaguidesteeringdiffusionpolices} and \citet{wang2025itps}, we also evaluate a diffusion policy (DP) released with open-source weights in previous work~\cite{du2025dynaguidesteeringdiffusionpolices}. The policies receive RGB observations from both a third-person camera and a wrist-mounted camera along with proprioceptive robot state information. At each replan step, \(\pi_{0.5}\) and DP predict an action chunk of future actions, from which the first $N$ actions are executed before replanning. We provide additional details on the simulation and real-world evaluation environments on the project webpage

\subsection{Shared Autonomy Arbitration}
Following classical policy-blending formulations~\cite{dragan2013policy}, we represent shared autonomy as arbitration between human and autonomous actions. In SAPS, the autonomous action is produced by a pretrained VLA, and arbitration is applied only after policy inference. Let:
\[\mathbf{a}_{\text{\textit{human}}},~\mathbf{a}_{\text{\textit{VLA}}} \in \mathbb{R}^7,\]
denote the human and policy actions, respectively, where the first 6 dimensions correspond to translational and rotational (3 DoF axis-angle representation) end-effector motion and the final dimension controls the gripper state. The final blended action executed is computed as:
\[\mathbf{a}^{(1:6)}_{\text{\textit{blended}}} = \alpha  \mathbf{a}^{(1:6)}_{\text{\textit{VLA}}} + (1 - \alpha)  \mathbf{a}^{(1:6)}_{\text{\textit{human}}},\]
where $\alpha \in [0, 1]$ determines the degree of autonomy. When no human input is detected, corresponding to the $\mathcal{L}_2$-norm of the human action falling below a threshold $\epsilon = 10^{-3}$, the robot policy retains full control with $\alpha = 1$. 

The gripper state \(a^{(7)}\) is handled separately. The user is able to toggle from the current state of the gripper, and the policy is not able to change from the user-initiated gripper state until the policy toggles gripper state twice in simulation, or a 5-second timeout occurs on hardware. These latching behaviors are designed to give the user more control over grasping when needed while reducing accidental drops. 

\subsubsection{Takeover Arbitration}
Takeover arbitration allows the human operator to completely override the VLA policy whenever active intervention is detected. Similar to blending-based methods, intervention is determined using the magnitude of the human command:
\[
\alpha =
\begin{cases}
0.0, & \left\lVert \mathbf{a}_{\text{\textit{human}}}^{(1:6)} \right\rVert > \epsilon, \\
1.0, & \left\lVert \mathbf{a}_{\text{\textit{human}}}^{(1:6)} \right\rVert \leq \epsilon.
\end{cases}
\]
where $\epsilon = 10^{-3}$ is used as the small activity threshold.

Unlike continuous blending approaches, Takeover uses switching between autonomous and human control, allowing the operator to fully correct policy behavior during execution. 

\subsubsection{Equal Blending Between Policy and Teleoperation}

Blending performs continuous arbitration between the VLA policy and human teleoperation, allowing shared autonomy. During active human intervention, a fixed arbitration coefficient of \(\alpha = 0.5\) is used, producing equal contribution from both the VLA and human. When no human input is detected, the controller transitions to full autonomy with \(\alpha = 1.0\).

Detection of intervention uses the magnitude of the human command, such that 
\[
\alpha =
\begin{cases}
0.5, & \left\lVert \mathbf{a}_{\text{\textit{human}}}^{(1:6)} \right\rVert > \epsilon, \\
1.0, & \left\lVert \mathbf{a}_{\text{\textit{human}}}^{(1:6)} \right\rVert \leq \epsilon.
\end{cases}
\]

\subsubsection{Cosine Similarity Agreement-Based Blending}

This method uses a dynamic $\alpha$, rather than a fixed blending coefficient, to adapt the autonomy level based on policy--human agreement. Cosine-based directional agreement between human and autonomous commands has previously been used for shared-control intent inference~\cite{jain2018recursive}. Here, we use this agreement directly to determine the action-level arbitration coefficient. At each step, we measure directional agreement \(c\) via Cosine similarity:
\[
c = \cos(\theta) = \frac{\mathbf{a}_{\text{\textit{human}}}^{(1:6)} \cdot \mathbf{a}_{\text{\textit{VLA}}}^{(1:6)}}{\lVert \mathbf{a}_{\text{\textit{human}}}^{(1:6)} \rVert , \lVert \mathbf{a}_{\text{\textit{VLA}}}^{(1:6)} \rVert}.
\]
The Cosine similarity is assigned an agreement score $g \in [0,1]$ through a logistic transformation that increases sensitivity near regions of disagreement while saturating for highly aligned actions: 
\[
\gamma = \sigma(k \cos(\theta)) = \frac{1}{1 + e^{-k \cos(\theta)}}, \qquad k = 6.
\]
The agreement score determines the arbitration coefficient, \(\alpha = \gamma\). 
This formulation enables smooth, continuous transitions between human and autonomous control, allowing corrective intervention while preserving stable behavior. This formulation also allows for confidence-based metrics \(c\) to replace \(\cos (\theta\)) in future work for arbitration. 





\section{User Study with Able-bodied Users}

\begin{figure}[tp]
    \centering
    \subfloat[User study setup.]{%
        \includegraphics[width=0.49\linewidth]{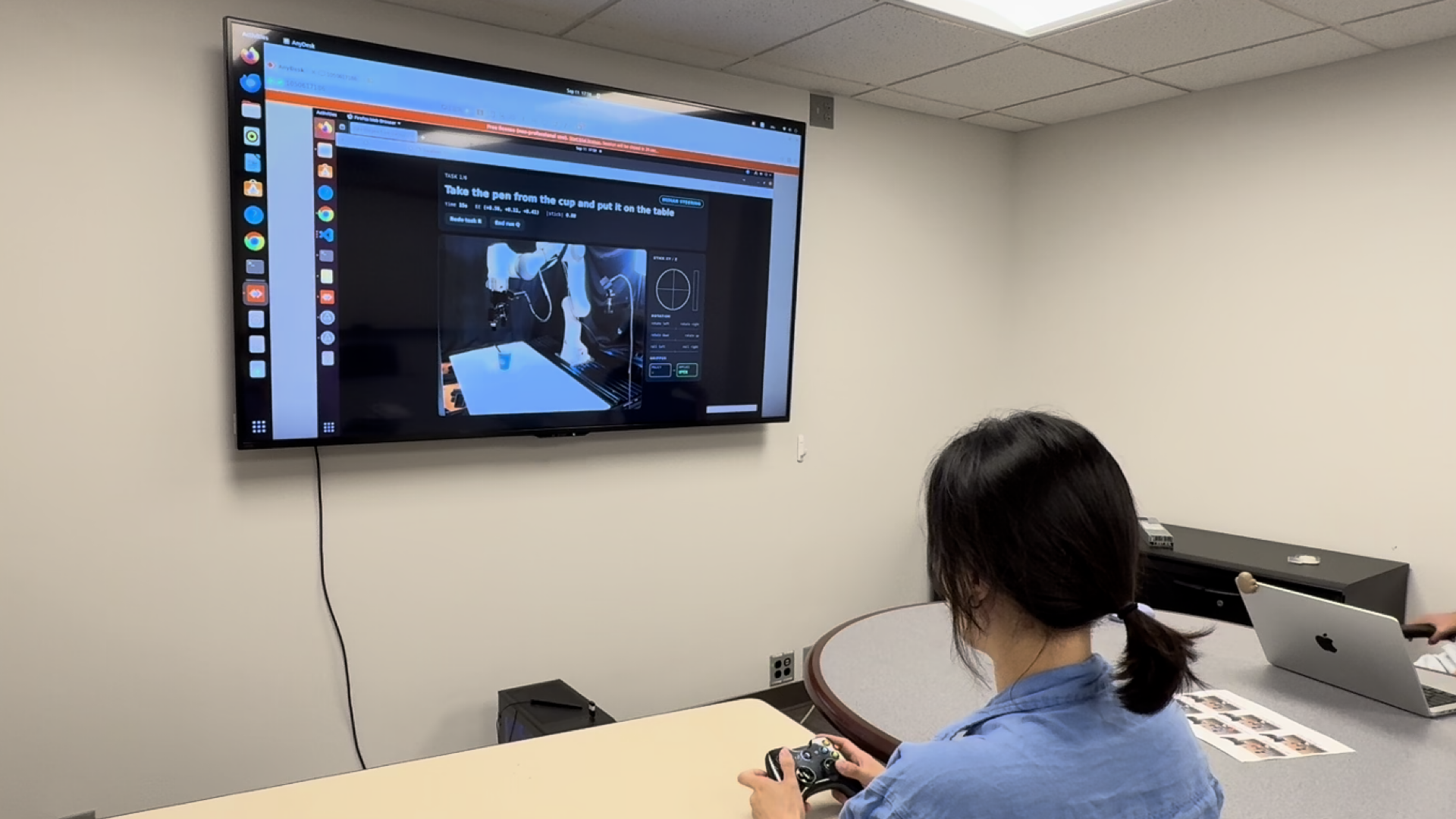}%
        \label{fig:user_study_setup}%
    }\hfill%
    \subfloat[Case study setup.]{%
        \includegraphics[width=0.49\linewidth]{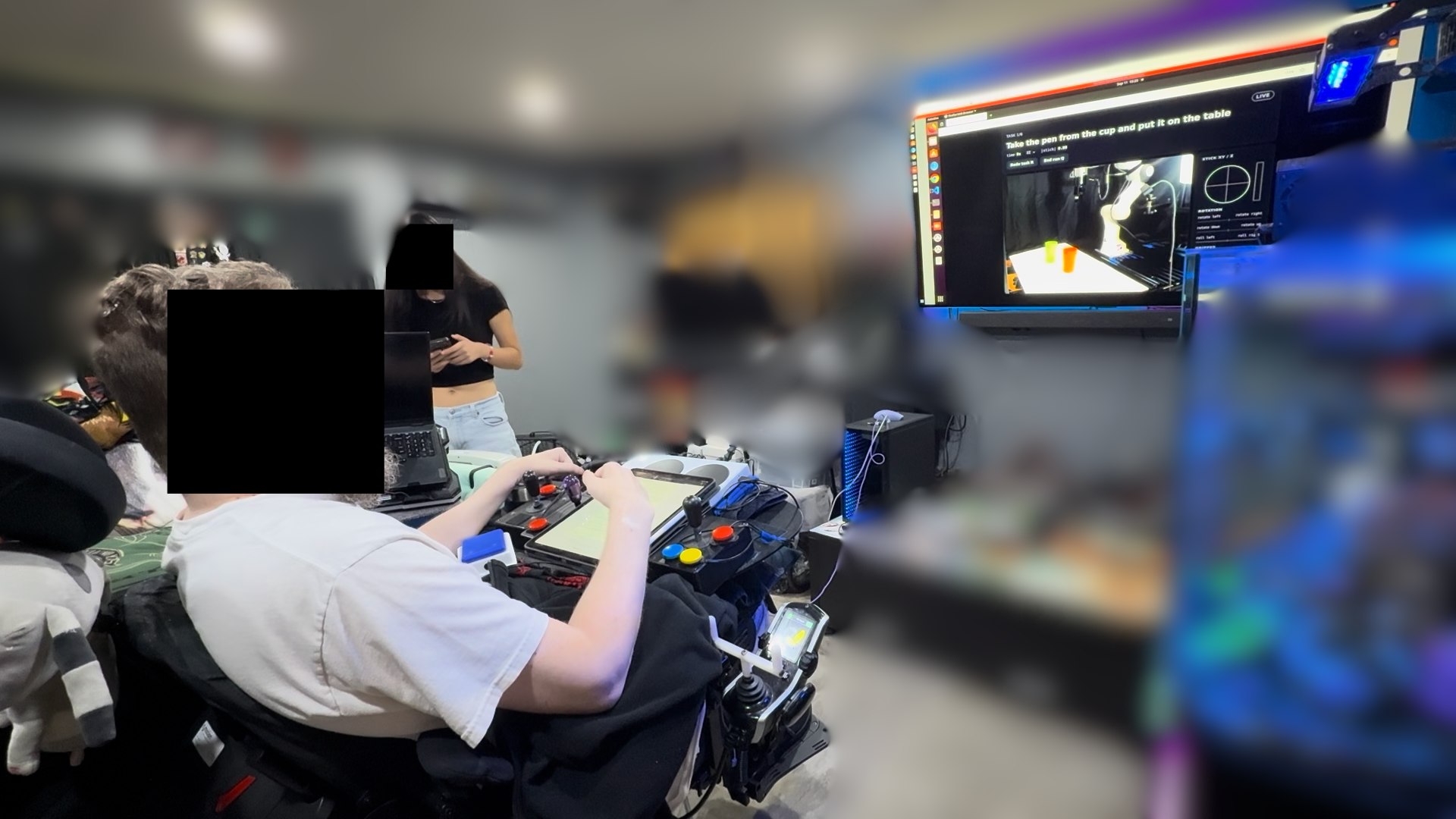}%
        \label{fig:case_study_setup}%
    }
    \caption{User study (\textit{N=14}) (left) and case study (right) setup. Both the user study and case study involve users controlling a Franka robot arm in simulation and the real world. The user study is performed in an office space, while the case study is performed in the user's home. }
    \label{fig:user_case_study_setup}
    \vspace{-1em}
\end{figure}

We run a user study to evaluate whether SAPS improves the performance of teleoperation and full autonomy. We test with 14 able-bodied users (ethics board approved, age 18-37, 9 male, 5 female) recruited from the local community through flyers. We test with two tasks in LIBERO-PRO and two tasks on hardware in the real world. To match the remote-teleoperation setting used in LIBERO, real-world participants viewed the task through a Logitech C920 webcam stream, as seen in Fig.~\ref{fig:user_study_setup}. Users reported median scores of 4.5 (IQR 2) for amount of experience with a joystick and 4 (IQR 3) for amount of experience controlling robots when asked 7-point Likert items rating experience from ``No Experience'' to ``Very High Experience.'' The active arbitration method was also hidden from users using a single-blind study design. 

LIBERO evaluations run a real-time 20 Hz control loop with asynchronous VLA replanning (3 Hz, 10 step action chunk) and per-step arbitration against a continuously-updated human input with 30-second timeouts. We use the default LIBERO normalized action space for blending, which when unnormalized becomes at max using translation in units of 0.05 m and rotation in units of 0.5 rad per time step. In the real world, we run a real-time 15 Hz control loop with synchronous VLA replanning by default (2 Hz, 8-step action chunk) and per-step arbitration against a continuously-updated human input with 50-second timeouts. Measured latency from controller to the real robot is around 15 ms. We tune the controller empirically and use a normalized action space for blending, which when unnormalized becomes at max using translation in units of 0.05 m and using rotation in units of 0.15 rad per time step. For real world tasks, we end tasks immediately in which an object is knocked over. 

Each task is repeated 3 times for the Teleoperation, Blending, Takeover, and Cosine methods resulting in a cumulative 12 episodes per task. Before each new environment, participants are given 3 minutes to warm up and become familiar with the environment view and gamepad controls. All tasks, environments, and arbitration sequences are counterbalanced to compensate for any training effects that occur. The simulation tasks include putting cups onto plates and putting objects into drawers. The real world tasks include stacking cups onto each other and taking a pen out of a cup. 

\subsection{Quantitative Results from User Study}
\label{subsec: user_study_quantitative_results}
Fourteen able-bodied operators completed LIBERO-PRO and thirteen operators completed the real world benchmark (one user encountered hardware setup difficulties). The setup is visualized in Fig.~\ref{fig:user_study_setup}. For all participant-level paired SAPS and Teleoperation comparisons, we use Wilcoxon signed-rank tests with Holm correction. Because Full-Autonomy runs are not able to be paired with individual users, we use Fisher's exact test for the unpaired Full-Autonomy baseline success rate comparison. Results from the user study in Table \ref{tab:real_world_study_results} indicate that SAPS can improve performance for users spanning a broad range of prior interface and robot-control experience.

Across SAPS configurations, task success was significantly higher than full autonomy in both simulation and hardware (all 6 pairwise Fisher's exact tests, Holm-corrected $p < 0.005$). Compared to teleoperation, SAPS task completion time differed significantly across methods in both simulation and hardware (\(p\le0.029\), \(p<0.001\) respectively; Friedman test). SAPS also significantly reduced active human intervention (all 6 pairwise Wilcoxon, Holm-corrected \(p<0.001\)) across benchmarks. Intervention was required for less than half of task time throughout the study. SAPS also significantly improved task success over teleoperation on hardware (all 3 pairwise comparisons on hardware, \(p\le0.017\); Wilcoxon). In simulation, only Cosine significantly improved success. Takeover was numerically higher but did not reach significance, while Blending had the same success rate as Teleoperation (Cosine: \(p=0.015\), Takeover: \(p=0.067\), Blending: \(p=0.92\); Wilcoxon). These quantitative improvements motivate examining whether the reduced control requirements are also reflected in users' perceived workload and preference.

\begin{table}[t]
    \centering
    \tiny
    \setlength{\tabcolsep}{3pt}
    \caption{Performance across baseline and SAPS methods in simulation (n = 14) and the real world (n = 13). Task completion time is computed over successful trials and intervention rate is reported as the percentage of time under human control.}
    \label{tab:study_results}
    \begin{tabular}{lccccc}
        \toprule
        & \multicolumn{2}{c}{Baselines}
        & \multicolumn{3}{c}{SAPS} \\
        \cmidrule(lr){2-3}
        \cmidrule(lr){4-6}
        \textbf{Metric}
        & \textbf{Full Autonomy}
        & \textbf{Teleoperation}
        & \textbf{Cosine}
        & \textbf{Blending}
        & \textbf{Takeover} \\
        \midrule

        \multicolumn{6}{l}{\textbf{Simulation (n = 14).}\label{tab:simulation_study_results}} \\
        Success Rate (\%)
        & 8.3
        & 44.0
        & \textbf{75.0}
        & 44.0
        & 65.5 \\

        Completion Time (s)
        & \textbf{13.3}
        & 23.2 \(\pm\) 3.1
        & 20.3 \(\pm\) 5.2
        & \textbf{17.8} \(\pm\) \textbf{4.2}
        & 21.4 \(\pm\) 5.3 \\

        Active Intervention (\%)
        & 0.0
        & 100.0
        & 42.9 \(\pm\) 14.2
        & 46.5 \(\pm\) 20.6
        & \textbf{32.2} \(\pm\) \textbf{10.3} \\

        \midrule
        \multicolumn{6}{l}{\textbf{Hardware (n = 13).}\label{tab:real_world_study_results}} \\
        Success Rate (\%)
        & 8.3
        & 34.5
        & 67.2
        & 64.3
        & \textbf{77.4} \\

        Completion Time (s)
        & 44.8
        & 42.4 \(\pm\) 4.5
        & 36.1 \(\pm\) 7.9
        & \textbf{34.4} \(\pm\) 6.9
        & 37.8 \(\pm\) 7.3 \\

        Active Intervention (\%)
        & 0.0
        & 100.0
        & \textbf{43.0} \(\pm\) \textbf{12.4}
        & 45.4 \(\pm\) 13.8
        & 44.3 \(\pm\) 10.2 \\

        \bottomrule
        \vspace{-3em}
    \end{tabular}
\end{table}

\subsection{Subjective Results from User Study}

We report subjective workload (NASA-TLX, 0--100; lower is better) and single-blinded method preference. Both were analyzed with Friedman tests across the four methods, followed by two-sided Holm-corrected Wilcoxon signed-rank tests comparing Teleoperation with each arbitration method.

\subsubsection{Workload}

\begin{figure}[t]\centering
  \includegraphics[width=0.7\linewidth]{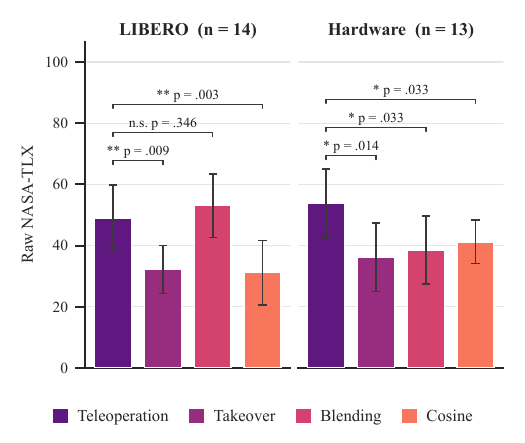}
    \vspace{-1em}
  \caption{Raw NASA-TLX by arbitration method. Bars are means with 95\% CI. Brackets compare teleoperation with each arbitration method  (Wilcoxon Signed-Rank Test, Holm-corrected).}
  \label{fig:tlx}
\end{figure}

\begin{table}[tp]
\centering
\vspace{-1em}

\caption{Raw NASA-TLX, $M$ ($SD$) for user study.}

\begin{tabular}{lcc}
\toprule
Method & LIBERO ($n{=}14$) & Hardware ($n{=}13$) \\
\midrule
Teleoperation & 49.0 (18.6) & 53.8 (18.7) \\
Takeover      & 32.2 (13.6) & 36.2 (18.6) \\
Blending      & 53.1 (18.0) & 38.5 (18.4) \\
Cosine        & 31.1 (18.2) & 41.2 (11.7) \\
\bottomrule
\label{tab:nasa-tlx}
\end{tabular}
\vspace{-2em}

\end{table}

Workload differed across methods on both benchmarks (LIBERO: $p<.001$; hardware: $p=.007$; Friedman). Table~\ref{tab:nasa-tlx} reports means and standard deviations; Fig.~\ref{fig:tlx} reports p-values, with positive $\Delta$ indicating lower workload than Teleoperation. In LIBERO, workload was lower for Cosine ($\Delta=17.9$) and Takeover ($\Delta=16.8$), but not Blending ($\Delta=-4.1$), likely because one task required sustained motion antiparallel to the VLA, limiting fixed $\alpha=0.5$ blending. On hardware, all three methods reduced workload: Takeover ($\Delta=17.6$), Blending ($\Delta=15.3$), and Cosine ($\Delta=12.6$).

\subsubsection{Preference}

\begin{figure}[t]\centering
  \includegraphics[width=0.7\linewidth]{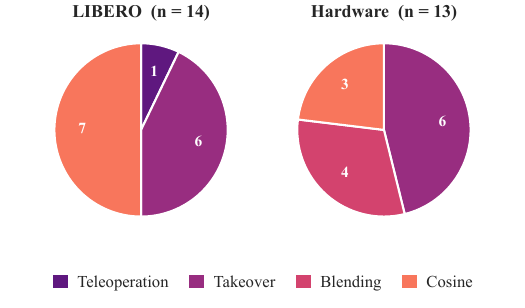}
    \vspace{-1.5em}
  \caption{First-place votes from the end-of-block forced ranking.}
  \vspace{-1.5em}
  \label{fig:pref}
\end{figure}

Single-blind forced ranking differed between methods on both benchmarks (LIBERO: $p<.001$; hardware: $p=.036$; Friedman test). In LIBERO, Cosine received 7 of 14 first-place votes and Takeover 6. Only Cosine ranked significantly above Teleoperation ($p=.013$). On hardware, Takeover received 6 of 13 first-place votes, followed by Blending (4) and Cosine (3), while Teleoperation received zero; only Takeover ranked significantly above Teleoperation ($p=.037$).

\section{Expert User Experiments}
\label{sec:result}

To evaluate additional tasks, benchmarks, and baseline methods, we conduct evaluations across LIBERO, LIBERO-PRO, CALVIN, and real-world robot tasks with 2 expert users. The expert evaluations complement the user study by enabling broader evaluation across tasks, benchmarks, and baselines that would be impractical to exhaustively test with participants, while the user study evaluated whether SAPS benefits users with varied robot-control experience.

\subsection{LIBERO}

\begin{figure}[b]

    \centering
    \vspace{-2em}
    \includegraphics[width=0.6\linewidth]{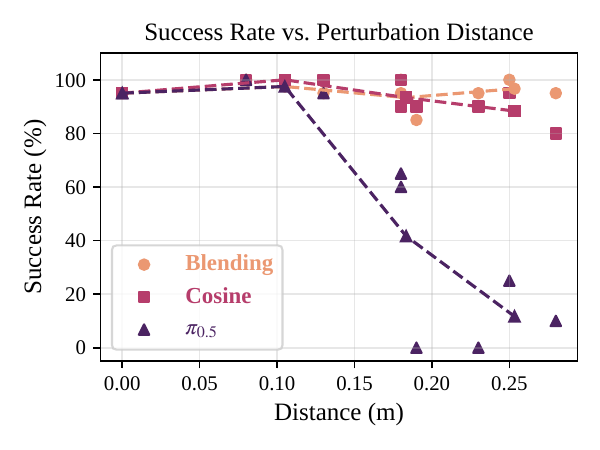}
    \vspace{-1.5em}
    \caption{\pizerofive{} task success rate degrades across 8 perturbed task setups (n = 20) in LIBERO as the target object’s displacement from its original position increases. Both Blending and Cosine maintain high success across larger displacements.}
    \label{fig:perturb-success}
\end{figure}

For the standard LIBERO evaluation~\citep{liu2023libero}, we use a controlled perturbation study on a single pick-and-place task from the LIBERO object suite: picking up the cream cheese and placing it in the basket. The position of the target object is systematically shifted away from the original training-distribution layout, illustrated on the project webpage. These offsets create out-of-distribution spatial configurations while preserving the same object identity, language instruction, and manipulation objective. Blending and Cosine are evaluated on the same perturbed task configurations as \pizerofive{}, using the gamepad interface for both Blending and Cosine. This experiment is intended to characterize the effect of shared human guidance under increasing spatial perturbation rather than to directly compare arbitration strategies.

\pizerofive{} performs well during testing when the target object remains close to the original task layout, whereas \(\pi_{0.5}\)'s performance decreases as the perturbation distance increases. Exact perturbation details can be found on the project webpage. This suggests that the policy retains useful manipulation behavior but is sensitive to spatial shifts in object placement. 

As shown in Fig.~\ref{fig:perturb-success}, \pizerofive{} success strongly decreases with perturbation distance. This trend is most visible at larger perturbations: for distances greater than or equal to $0.15$ m, \(\pi_{0.5}\) drops to $22.9\%$ success, while SAPS Blending and Cosine configurations remain on average at $94.4\%$ and $93.3\%$ success, respectively. Overall, Blending and Cosine maintain high success rates across perturbation distances, showing that action-level shared autonomy can recover from many spatial generalization failures as the autonomous policy becomes less reliable. This controlled study motivates the broader LIBERO-PRO evaluation, where SAPS is evaluated across structured task and scene perturbations.

\subsubsection*{LIBERO-PRO}
\label{subsec: LIBERO-PRO}
\begin{figure}[t]
    \vspace{-1.5em}
    \centering
    \includegraphics[width=0.6\linewidth]{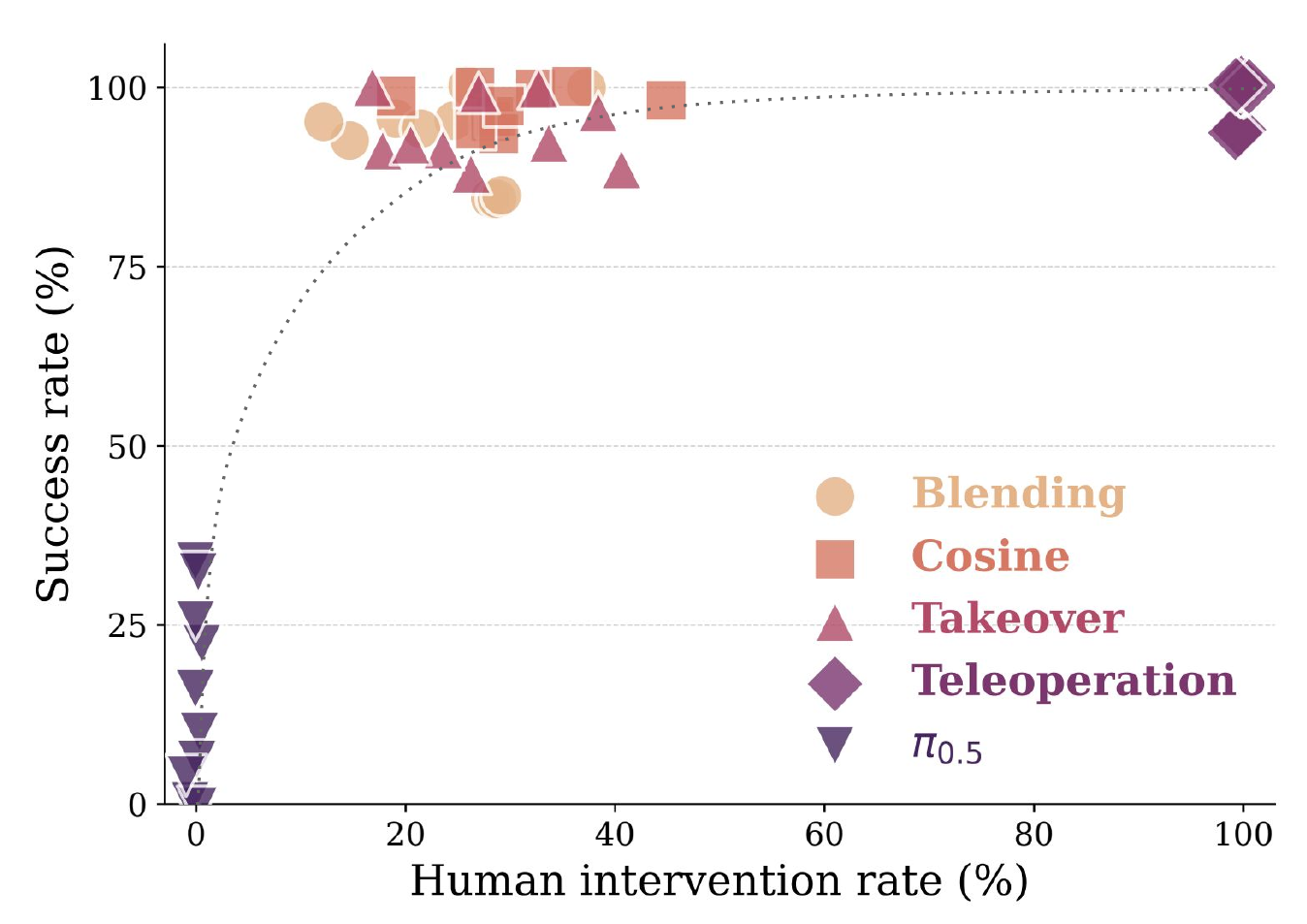}
    \vspace{-1em}
    \caption{With SAPS methods, we find in LIBERO that similar success rates to pure Teleoperation are achieved for a fraction of the effort from an operator. The dashed trendline summarizes that a small fraction of human intervention produces large gains in performance over \pizerofive{}.}
    \label{fig:libero_pro_intervention_vs_sr}
    \vspace{-1.5em}
\end{figure}

LIBERO-PRO extends the LIBERO with 4 perturbation variations for each task~\citep{zhou2025liberopro}. In this work, we select 10 tasks for an expert user to evaluate SAPS and focus on two perturbations: task and swap. We choose these because they introduce the largest changes to object identity, placement, and task specification. Task perturbations alter the language-conditioned goal, requiring the policy to bind the correct object and instruction. Swap perturbations change the positions of relevant objects with other objects present in the scene, requiring the policy to recover from spatial shifts. The selected tasks include long-horizon, pick-and-place, and mug manipulation tasks. More information on selected tasks is found on the project webpage.

We use LIBERO-PRO to evaluate whether SAPS improves zero-shot robustness under structured task and swap perturbations. Across the 10 selected tasks, autonomous \pizerofive{} achieves only $15.0\%$ mean success, indicating that the policy is highly sensitive to out-of-distribution changes in object layout and task specification. In contrast, all SAPS arbitration strategies achieve \(92.6\%\)--\(97.4\%\) mean success, approaching Teleoperation’s $98.8\%$ success rate. The results are consistent with \pizerofive{} retaining useful manipulation behavior, but human guidance being needed to maintain correct task behavior under perturbations.

\begin{figure}[htp]
    \vspace{-1em}
    \centering
    \includegraphics[width=0.65\linewidth]{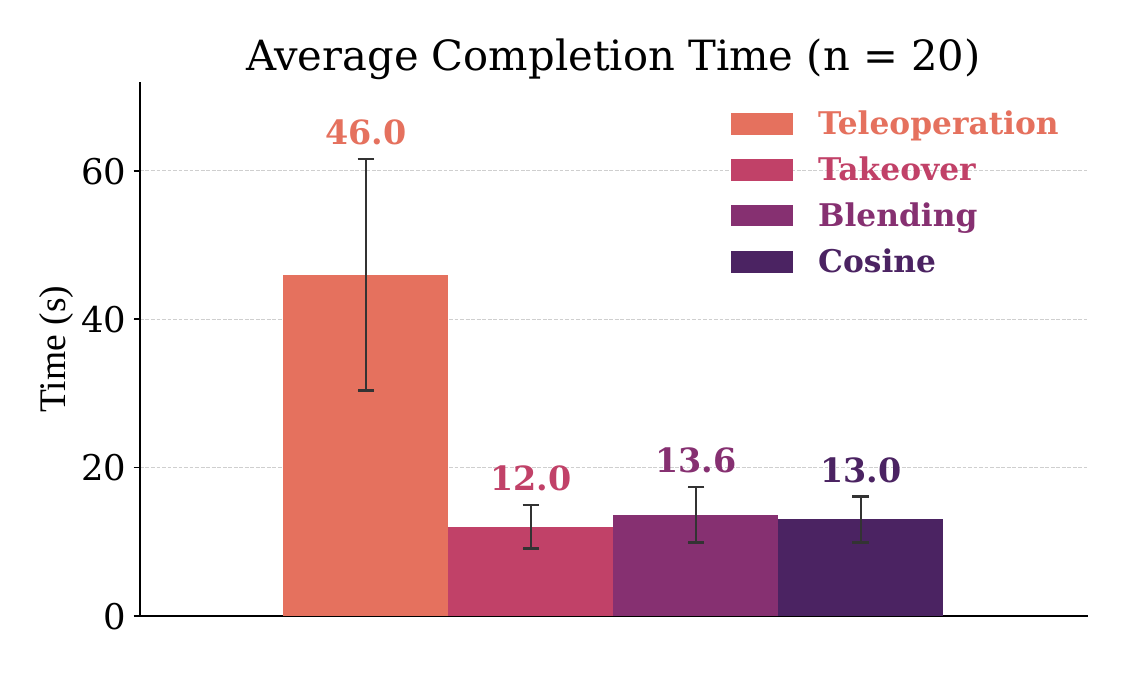}
    \vspace{-1em}
    \caption{Completion time across LIBERO-PRO simulation tasks for \pizerofive{}, Teleoperation, Takeover, Blending, and Cosine. Shared-autonomy methods reduce completion time relative to \(\pi_{0.5}\) and Teleoperation across all tasks with $p<0.006$.}
    \label{fig:libero-pro_completion_time}
    \vspace{-1em}
\end{figure}

Completion-time results of successfully completed tasks in Fig.~\ref{fig:libero-pro_completion_time} show that this improvement does not come at the cost of greater completion time or continuous manual control. \pizerofive{} often enters incorrect or inefficient recovery behaviors, while the expert user must control the entire task during Teleoperation with limited depth perception in simulation. SAPS avoids both failure modes: human input guides the system back toward familiar states from which the policy can continue the task. Across arbitration strategies, SAPS completes successful trials in \(11.1{-}13.2\)s on average, compared with \(30.7\)s for autonomous \pizerofive{} and \(46.0\)s for Teleoperation. We evaluate the results with Friedman test (p=0.0004) followed by two-sided Wilcoxon paired by task and Holm-correction. All three shared-autonomy methods are significantly faster than teleoperation (\(p<0.006\); Wilcoxon paired across tasks for all 3 SAPS configurations). Because \pizerofive{} completed 0 tasks successfully for 3 of the 10 tasks, Wilcoxon paired tests across tasks are not run for task times comparing SAPS with \pizerofive{}. 

Human intervention rates show that SAPS improves performance without reverting to continuous manual control. Across LIBERO-PRO tasks, SAPS requires active human input only \(10.8{-}30.0\%\) of steps, compared with \(100\%\) in Teleoperation.


\subsection{CALVIN}

CALVIN evaluates language-conditioned manipulation in a Franka tabletop kitchen environment. In the long-horizon \pizerofive{} setting, all 34 CALVIN skills are composed into five-subtask chains with persistent environment state, so failures or imprecise motions can carry over to later subtasks. This chaining procedure is performed in the original CALVIN benchmark~\citep{mees2022calvin}. In CALVIN, we run a real-time 30 Hz control loop with synchronous VLA replanning (6 Hz, 10-step action chunk) and per-step arbitration against a continuously-updated human input with 12-second timeouts. We use the default CALVIN normalized action space for blending, which when unnormalized becomes at max using translation in units of 0.02 m and using rotation in units of 0.05 rad per time step.


We evaluate SAPS with \pizerofive{}~\cite{pi05}, trained on CALVIN ABC and tested on CALVIN D, using CALVIN's standard task evaluator. 
For CALVIN, the unguided \pizerofive{} baseline is strong, completing $91.27\%$ of attempted subtasks 
and all five subtasks in $63.33\%$ of chains due to the fast timeout of CALVIN. Because the standard CALVIN timeout strongly constrains manual control, we additionally evaluate Teleoperation with a \(3\times\) timeout, increasing subtask success from \(47.37\%\) to \(86.78\%\).
SAPS outperforms both autonomous \pizerofive{} and Teleoperation, reaching \(94.85\%\) subtask success and a \(76.67\%\) five-in-a-row success rate while requiring human input on only \(13.7\%\) of environment steps with the positive gap in success increasing as the horizon gets longer.


\begin{table}[b]

  \centering
  \tiny
   \vspace{-3em}
  \setlength{\tabcolsep}{4pt}
  \caption{Results for \pizerofive{} policy steering on CALVIN ($n{=}30$ five-subtask chains). ST-SR denotes overall subtask success, Mean/5 the mean subtasks per chain, ST-X the success rate at subtask $X$, Human \% percent of steps with user input, and T/O timeout. }

  \begin{tabular}{l|cc|ccccc}
    \toprule
    Method & ST-SR (\(\uparrow\)) & Mean / 5 (\(\uparrow\)) & ST-1 (\(\uparrow\)) & ST-2 (\(\uparrow\)) & ST-3 (\(\uparrow\)) & ST-4 (\(\uparrow\)) & ST-5 (\(\uparrow\)) \\
    \midrule
    \pizerofive{} & 91.27\% & 3.833 & 90.00\% & 83.33\% & 73.33\% & 73.33\% & 63.33\% \\
    Teleoperation (1x T/O) & 47.37\% & 0.900 & 53.33\% & 26.67\% & 6.67\% & 3.33\% & 0.00\% \\
    Teleoperation (3x T/O) & 86.78\% & 3.500 & 93.33\% & 80.00\% & 76.67\% & 53.33\% & 46.67\% \\
    Cosine & \textbf{94.85\%} & \textbf{4.300} & \textbf{96.67\%} & \textbf{90.00\%} & \textbf{83.33\%} & \textbf{83.33\%} & \textbf{76.67\%} \\
    \bottomrule
  \end{tabular}
  \hfil
  \begin{tabular}{l|cc|c}
    \addlinespace[0.5em]
    Method & EE Path (m) (\(\downarrow\)) & Timesteps (\(\downarrow\)) & Human \% \\
    \midrule
    \pizerofive{} & 0.717 & 116.8 & 0.00\% \\
    Teleoperation (1x T/O) & 1.809 & 299.7 & 100.00\% \\
    Teleoperation (3x T/O) & 1.182 & 484.8 & 100.00\% \\
    Cosine & \textbf{0.615} & \textbf{112.1} & 13.73\% \\
    \bottomrule
  \end{tabular}

  \label{tab:calvin_chain_n30}
\end{table}


To compare SAPS with two state-of-the-art policy steering methods, DynaGuide and ITPS, we additionally evaluate Cosine on a lower-parameter DP baseline for independent single-subtask evaluations, following the same evaluation performed in~\citet{du2025dynaguidesteeringdiffusionpolices}. 
The unguided DP baseline executes autonomously, while ITPS~\cite{wang2025itps} steers diffusion sampling with a human-specified position-target gradient and DynaGuide~\cite{du2025dynaguidesteeringdiffusionpolices} uses an auxiliary dynamics model to guide sampling toward desired future states. 
Despite the simpler action-level arbitration formulation, SAPS shared autonomy outperforms the autonomous DP baseline, ITPS, and DynaGuide on all 34 CALVIN subtasks over 10 episodes each, averaging 93\% success compared to 45\%, 66\%, and 80\%, respectively. 
Additional details, including the results on individual subtask success rates, per-method steering mechanisms, intervention rates, end-effector path length, completion time, and qualitative rollouts, are also provided in the project webpage.

\subsection{Real World}
\label{subsec: real_world_expert}

Real-world expert-user evaluations use a Franka Emika Panda arm controlled by \pizerofive{} finetuned on the DROID dataset. Unlike in the user study, we evaluate three tabletop manipulation tasks with the user in the same room: picking up and placing the marker off the plate, closing a drawer, and opening the left door on a cabinet. These tasks cover object transport, contact-rich pushing, and articulated-object manipulation, expanding task types from the user study. Each task is evaluated across \(\pi_{0.5}\), Teleoperation, and SAPS characterized by Blending and Cosine arbitration.  

As seen in Table~\ref{tab:real_success}, $\pi_{0.5}$ struggles across real-world tasks, achieving 50\% success on Pick and Place, 25\% success on Close Drawer, and 5\% success on Open Cabinet (with task pictures shown on the project webpage). Shared autonomy substantially improves real-world reliability, with Cosine achieving an average success rate of $98.3\% $ across all three tasks, and Blending achieving $93.3\%$. Additional task-specific success results can be found in Table~\ref{tab:real_success}. This suggests that $\pi_{0.5}$ has learned useful manipulation behavior from DROID, but remains brittle under specific task layouts and object states, while corrective human input helps recover from these failures. 

\begin{table}[t]
    
    \centering
    \scriptsize
    \caption{Hardware evaluation ($n=20$) by task success rate.}
    \begin{tabular}{lcccc}
        \toprule
        Task & \pizerofive{} & 
        Teleoperation &
        \begin{tabular}{@{}c@{}}Blending\end{tabular} &
        \begin{tabular}{@{}c@{}}Cosine\end{tabular} \\
        \midrule
        Pick and Place & 50\% & 100\% & 100\% & 100\% \\
        Close the Drawer & 25\% & 100\% & 80\% & 100\% \\
        Open the Left Cabinet & 5\% & 100\% & 100\% & 95\%
        \\
        \bottomrule
    \end{tabular}
    \vspace{-2.0em}
    \label{tab:real_success}
\end{table}

Completion-time results further show that shared autonomy improves task efficiency relative to $\pi_{0.5}$. On average, Blending is faster by $99.6 s$ and Cosine is faster by $116.3s$  compared to \pizerofive{} across all hardware tasks. More detailed results can be found on the project webpage. Here, we use pairwise Mann--Whitney U tests for each task because this is evaluated across a single user across only 3 tasks. We find that both Cosine and Blending significantly reduce completion time relative to $\pi_{0.5}$ across all three tasks ($p \leq 0.0212$), indicating that shared autonomy not only improves success but also reduces inefficient autonomous recovery behavior. However, completion-time results compared to Teleoperation are task-dependent: both Cosine and Blending are significantly faster on the Marker Plate (Cosine: $p=0.0002$; Blending: $p=0.0005$), but neither method is significantly faster on Close Drawer or Open Cabinet. This differs from simulation, where Teleoperation is slower and the user lacks depth perception.


\section{Case Study: Expert User with Quadriplegia}
We conduct an assistive-teleoperation case study with a single user (age 40, male) with quadriplegia. The user has an AIS A C5 spinal cord injury, with the injury occurring 20 years ago. The setup is visualized in Fig.~\ref{fig:case_study_setup}. This user reported very high prior experience both with joystick usage and controlling robots (both 7 on 7-point Likert scales), and participates as a user of robot arms in research studies through the use of brain computer interfaces and other assistive interfaces. In this study, the user used an Xbox Adaptive Controller, which allowed him to use large joysticks and buttons using residual shoulder movements. 

Through SAPS arbitration methods, the user increased success rates and reduced task times by 8.3\% and 16.9\% compared to teleoperation, respectively in LIBERO. Using real-world hardware, the user reduced task times by 15.5\% compared to teleoperation but had an equal success rate. 

SAPS reduced workload with Takeover and Cosine in LIBERO for two methods (Teleoperation 39.2, Takeover \textbf{25.0}, Blending 68.3, Cosine \textbf{28.3}). However, we found that the user reported higher workload for their SAPS arbitration methods in hardware (Teleoperation \textbf{27.5}, Takeover 36.7, Blending 44.2, Cosine 40.0). The user-reported Teleoperation workload was much lower than that reported by other users, where hardware teleoperation workload was 49\% less than the mean of the user study. In contrast, mean hardware arbitration workload was reported as 4\% higher than the user study group mean. The user also ranked teleoperation as their most-preferred method through forced ranking for both LIBERO and hardware. This preference is consistent with findings in assistive shared autonomy that users may favor retaining greater control even when doing so does not optimize objective task performance~\cite{gopinath2017human}.

We believe the subjective results may be due to the user’s unusually high level of expertise in robot teleoperation, having reported themselves as ``Very Experienced'' in robot control. 
Despite the subjective preference for teleoperation, this motivates future work investigating whether additional training and familiarity with shared autonomy reduce workload for highly experienced teleoperation users~\cite{yang2026bimanualemg}. In our current work we give the user no training with SAPS and blind the user to the method used for all arbitration methods to reduce bias. 

\section{Lessons Learned}
\subsubsection{Operator experience shapes steering efficiency} Results in Sections \ref{subsec: LIBERO-PRO} and \ref{subsec: real_world_expert} are consistent with the user-study findings in Section \ref{subsec: user_study_quantitative_results}, showing that the performance benefits of SAPS persist across levels of operator experience. In LIBERO-PRO, experienced operators achieved 92.6--97.4\% success with SAPS, near the 98.8\% teleoperation ceiling, while requiring human input on only 10.8--30.0\% of execution. In the broader user study, participants required more intervention but improved success rates in shared autonomy versus teleoperation, suggesting that operator experience may affect how efficiently users can steer the policy, while less-experienced users may benefit more from shared autonomy. A similar trend appeared on hardware: SAPS preserved near-ceiling performance for expert operators while providing larger performance gains when direct teleoperation was more challenging for users in the user study.

\subsubsection{Objective performance does not capture user preference} The case study suggests that the value of shared autonomy depends on both objective performance and individual user preferences. The expert user with quadriplegia achieved comparable or improved task performance with SAPS, while preferring teleoperation and reporting lower workload with it on hardware. Together, these results motivate shared-autonomy systems that adapt to user expertise and preferences, and highlight the importance of evaluating performance, intervention, workload, and preference as complementary measures.

\section{Limitations}
SAPS requires a human operator at test time, so performance depends on operator timing, skill, and the teleoperation interface. Because SAPS uses blending with the policy rather than explicit task progress, intent, uncertainty, contact, or safety estimates, it can fail when the policy is confidently wrong. We alleviate this effect with Cosine similarity and Takeover, because the user can take over when the policy action differs from the user’s intended action. We also acknowledge that Cosine and Blending may cause objects to run into obstacles, e.g. when the policy is trying to go left around an obstacle while the user may want to go perpendicular to the policy by going to the right. In practice, users using Cosine can direct the robot to go back the way it came by moving antiparallel to where the policy is heading. However, more dynamic arbitration methods may be necessary in future work for more efficient shared autonomy in scenarios of partial disagreement. 
Our evaluation is limited to a finite set of simulated and real manipulation tasks; broader LIBERO coverage, more operators, additional embodiments, additional policy backbones, and more real-world tasks would better characterize improvement using SAPS.

\section{Conclusion}
\label{sec:conclusion}
We introduced SAPS, a lightweight shared-autonomy framework that steers pretrained robot policies and VLAs by blending policy actions with human teleoperation commands.
Across LIBERO, LIBERO-PRO, CALVIN, and real Franka hardware, SAPS improves robustness over autonomous execution while reducing human intervention relative to pure teleoperation. These results suggest that shared autonomy is a practical complement to robot foundation models. Coarse human input can guide policies through out-of-distribution failures while preserving their learned manipulation behavior. This improvement with shared autonomy is seen both in simulation environments and in the real world, where shared autonomy can reduce human effort and completion time for tasks requiring precise movements.
\bibliographystyle{IEEEtranN}
\bibliography{references}  

\end{document}